\documentclass[acmlarge,screen,nonacm]{acmart}
\usepackage{multirow}
\usepackage{booktabs}
\usepackage{tabularx,pbox}
\usepackage{makecell}

\AtBeginDocument{%
  \providecommand\BibTeX{{%
    \normalfont B\kern-0.5em{\scshape i\kern-0.25em b}\kern-0.8em\TeX}}}

\begin{document}

\title{Towards Integrative Multi-Modal Personal Health Navigation Systems: Framework and Application}

\author{Nitish Nag}
\authornote{These authors contributed equally to this research.}
\email{nitish.nag@zepp-usa.com}
\orcid{0000-0002-8409-1223}
\author{Hyungik Oh}
\authornotemark[1]
\email{jordan.oh@zepp-usa.com}
\author{Mengfan Tang}
\authornote{Corresponding author}
\email{mengfan.tang@zepp-usa.com}
\author{Mingshu Shi}
\author{Ramesh Jain}
\affiliation{%
  \institution{Zepp Health}
  \state{California}
  \country{USA}
}



\renewcommand{\shortauthors}{Nag and Oh, et al.}

\begin{abstract}

It is well understood that an individual's health trajectory is influenced by choices made in each moment, such as from lifestyle or medical decisions. With the advent of modern sensing technologies, individuals have more data and information about themselves than any other time in history. How can we use this data to make the best decisions to keep the health state optimal? We propose a generalized Personal Health Navigation (PHN) framework. PHN takes individuals towards their personal health goals through a system which perpetually digests data streams, estimates current health status, computes the best route through intermediate states utilizing personal models, and guides the best inputs that carry a user towards their goal. 

In addition to describing the general framework, we test the PHN system in two experiments within the field of cardiology. First, we prospectively test a knowledge-infused cardiovascular PHN system with a pilot clinical trial of 41 users. Second, we build a data-driven personalized model on cardiovascular exercise response variability on a smartwatch data-set of 33,269 real-world users. We conclude with critical challenges in health computing for PHN systems that require deep future investigation.
\end{abstract}




\maketitle

\section{Introduction and Related Work}
\begin{center}
“A record if it is to be useful to science, must be continuously extended, it must be stored, and \textit{above all it must be consulted} - Vannevar Bush \cite{bush1945we}”
\end{center}

The World Health Organization defines health “as a state of complete physical, mental and social well-being and not merely the absence of disease or infirmity” \cite{WorldHealthOrganization2005WHO2005}. How can one achieve this state of well-being? Health is a dynamic state that is changing as it relates to our unique biology, environment, and lifestyle. With modern technologies, users ubiquitously and continuously produce vast amounts of personal data while having unprecedented computing power at their fingertips. \textit{A serious challenge remains in capitalizing this opportunity of transforming data to impact real-world improvement in personal health.} We posit that a multi-modal (MM) detailed record that is continuously used to nurture personalized health using modern learning and cybernetic techniques is imperative. 

Since the origin of personal computing, large research efforts have been placed on data-driven personalization. In the 2000's the key challenges focused on harnessing data for personalizing social media and business applications \cite{Bennett2007ThePrize}. In the 2010's decade, machine learning and pattern recognition advancements allowed for more personalized services \cite{Weiss2012TheRecognition,Qumsiyeh2012PredictingRecommendations}. Success of large entities such as Google, Amazon, Facebook, Netflix were fueled by this multi-decade long effort on personalization \cite{Montgomery2009ProspectsInternet,Smith2017TwoAmazon.com}. The 2020's decade presents the health computing research community a nexus of continuing sensor advancements, disparate data, computing power, and great societal needs in health \cite{Boll2019HealthMedia19, Boll2018HealthInsights,Boll2018MMHealthCare}. \textbf{Effective personalized health computing is the next frontier of advancing well-being for people.} Below, we curate relevant contributions in digital health computing that have made progress in key elements of this challenge.

\textbf{Sensing:} Continuous sensors, digital interactions, and biological measurement devices produce vast amounts of data on an individual’s perception, biology, environment, and life events. Modern health data can be organized in two axes: 1) the data quality level and 2) the data collection frequency. Continuous digital sensors including computers, smartphones, and wearables, are relatively inexpensive in cost, energy, and computing power \cite{Fletcher2010WearableCare,Castignani2015DriverMonitoring,Feng2015CitizenSmartphones,Stupar2012WearableSensor}. The advancement in these technologies make it widely available for monitoring individual health data streams \cite{Majumder2019SmartphoneDiagnosis}. While continuous sensing gives us information immediately at the cost of quality (i.e. wearables), non-continuous methods are critical for providing high-resolution quality information (i.e. blood labs, medical imaging) to present a baseline understanding of an individual’s health status \cite{Springer, Kos2019WearableFeedback, Paradiso2005ASensors}. Various approaches have been developed for multi-sensor data acquisition and synchronization \cite{kaempchen2003data,raman2020modular,bannach2009automatic}. These aspects of sensor advancement create a foundation of personal data by which we can apply computing techniques towards personalization.

\textbf{Life-Logging:} Since Vannevar Bush applied a systematic approach to life-logging in 1945 \cite{bush1945we}, there have been many efforts on understanding personal experiences through the quantified self. One of the most representative researchers, Gordon Bell, used a wearable camera to capture a series of everyday life to aid recollection of past experiences \cite{gemmell2002mylifebits, gemmell2006mylifebits, bell2009total}. Gurrin et al, have been retrieving various semantic concepts through MM data streams using wearable devices and smartphones \cite{wang2016characterizing, gurrin2014lifelogging, albatal2013senseseer}. Event-indexed personal chronicle databases have been in active development through life-logging, event detection techniques to build personal models, and life-logs focused on mental health \cite{Oh2017FromChronicles, jalali2014personicle, mack2021mental, wang2014studentlife, jain2014objective}. However, to the best of our knowledge, there is no life-logging technique or tool that can compute appropriate actionable steps for health.


\textbf{Health State Estimation (HSE):} The Health State (HS) gives a current synopsis of a user’s health status through quantitative assessment in multi-dimensional space by fusing MM data and medical domain knowledge \cite{Nag2018Cross-modalEstimation, Muller2017RetrievalReview}. Development of techniques that combine low and high quality data streams to further improve accuracy of state estimation, such as Kalman filtering, GPS/INS integration and reinforcement learning are also an active part of work in this area \cite{el2019end,Musoff2009FundamentalsEdition, Qi2002DirectIntegration,Work2008AnDevices}. HSE research alone unfortunately lacks in producing actionable next steps once knowing the state, and how to use continuous state estimations in various contexts.

\textbf{Personal Modeling (PM):} By modeling an individual, a system can gain two personalization advantages. First, we can understand why and how the state of health evolves from a variety of inputs \cite{Tarassenko2018MonitoringPeople, Qu2015IndividualityCells, Arevalo2016ArrhythmiaModels}. This model can then create data-driven actions and predictions to improve health. Second, is modeling what causes certain actions to be taken. This requires pattern building of events pertaining to behavior or context, such as in lifestyle event mining \cite{RostamiPersonalModel, Nag2017HealthObservations}. Most MM computing models use group based methods such as collaborative filtering and deep learning models, are usually with shopping, advertising, and entertainment experiences where there is ample data available, interactions with users are not costly. This has severe limitations when addressing personal health models. For this reason, personal models require unique N=1 statistical methodology that is being developed \cite{Lillie2011TheMedicine,Guyatt1990TheExperience,McQuay1994DextromethorphanDesign,Odineal2020EffectTrial}.

\textbf{Health Recommender Systems:}
Current research in health recommendation engines generally provide either simplified automation or meld human expertise with computing aid for decision making. Automated systems are unfortunately limited in their function, mostly used to maintain simple homeostatic control, such as controlling glucose with an artificial pancreas \cite{Albisser1974ClinicalPancreas,Kovatchev2018AutomatedPancreas}. Human experts may be utilized for recommending inputs to an individual’s health state in-person or through telecommunication and decision aid systems \cite{Perednia1995TelemedicineApplications,Tuckson2017Telehealth,Al-Majeed2015HomeIoT}. These forms of guidance can be synchronous or asynchronous through various forms of communication and MM. Although efforts are being made to make these methods of recommendations better, the key limitation in the current recommender systems is the lack of sophisticated personalization, long-term intelligent planning, and contextual awareness \cite{Nag2017HealthObservations,Jovanov2015PreliminaryMonitoring,Ferrari2020OnRecognition,Baig2017AAdoption}.

A fundamental reason for the limitations of broad impact in the above areas of MM health computing is due to the narrow scope that each component addresses towards the overarching goal of keeping people in a healthy state. Our motivation is this necessity to bring these disparate computing efforts together in a user-centered and synergistic manner.

\textit{The key contribution of this work is both the generalized fusion of the aforementioned isolated components into a complete and general computing system. With real-world testing we demonstrate it's efficacy with a combined team of experts in multi-modal computing, medical sciences, and global-scale industry. This framework, which we call Personal Health Navigation (PHN) creates a computing environment to perpetually meld MM sensing to estimate health states continuously, build personal models to understand an individual, and provide actionable guidance to help individuals/healthcare professionals towards desired health goals. It is the first time a general computing system framework for health navigation has been proposed for processing MM data beyond a concept \cite{Nag2019ALife}. Finally, PHN opens a multitude of health computing research challenges that have substantial impact.}

\begin{figure}
\small
\centering
\includegraphics[width=.7 \columnwidth]{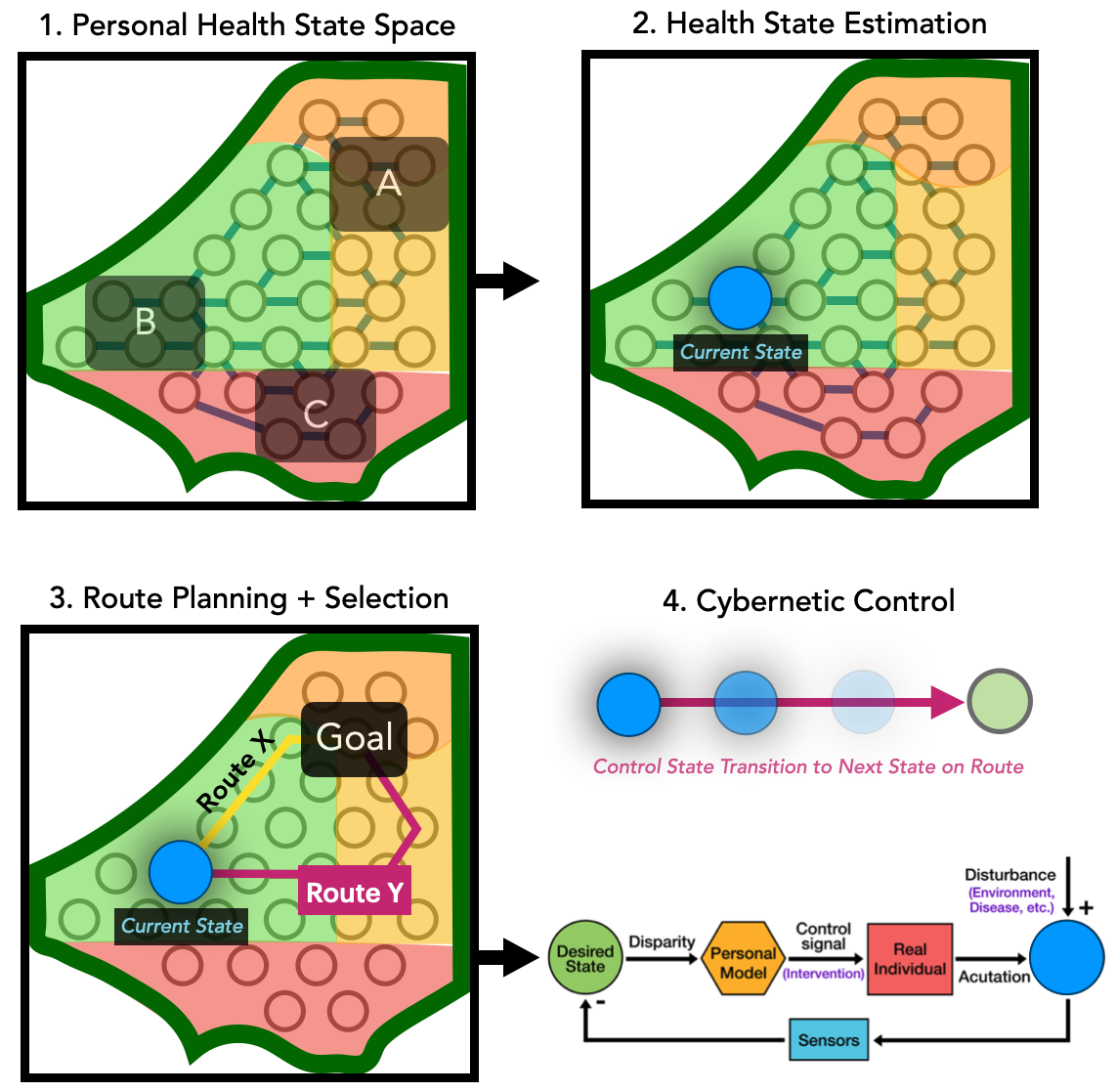}
\caption{Personal Health Navigation Concept: 1) We establish the state space for an individual. 2) HSE provides the user with current status of health. 3) Given a goal, the user is provided route options to select in order to reach the goal. 4) The user is guided through cybernetic control to maintain inputs to arrive at the next neighboring state on the route. Step 3 is performed again for re-planning when the state is updated and this cycle between steps 3 and 4 continues till the user reaches and maintains the goal state.}~\label{fig:phnconcept}
  \vspace{-6mm}
\end{figure}

\section{General PHN System Framework}
Navigation is a goal-based guidance system that routes a user through intermediate states by providing contextually relevant guidance. Through the combination of sensors, estimation techniques, intelligent planning, control systems, actuation mechanisms, and goal-decomposition, each iteration of navigation attempts to bring the user closer to the goal. When deployed in reality, the system must effectively cope with stochastic, chaotic, and noisy environments. The system engages in this goal-driven loop of sensing, computing, and actuating inputs by combining the pioneering elements from cybernetic systems for control and intelligence techniques to understanding the relevant state space \cite{Newell2013GPSThought,Wiener1962CyberneticsMachine}.

\begin{figure}
\small
\centering
\includegraphics[width=1 \columnwidth]{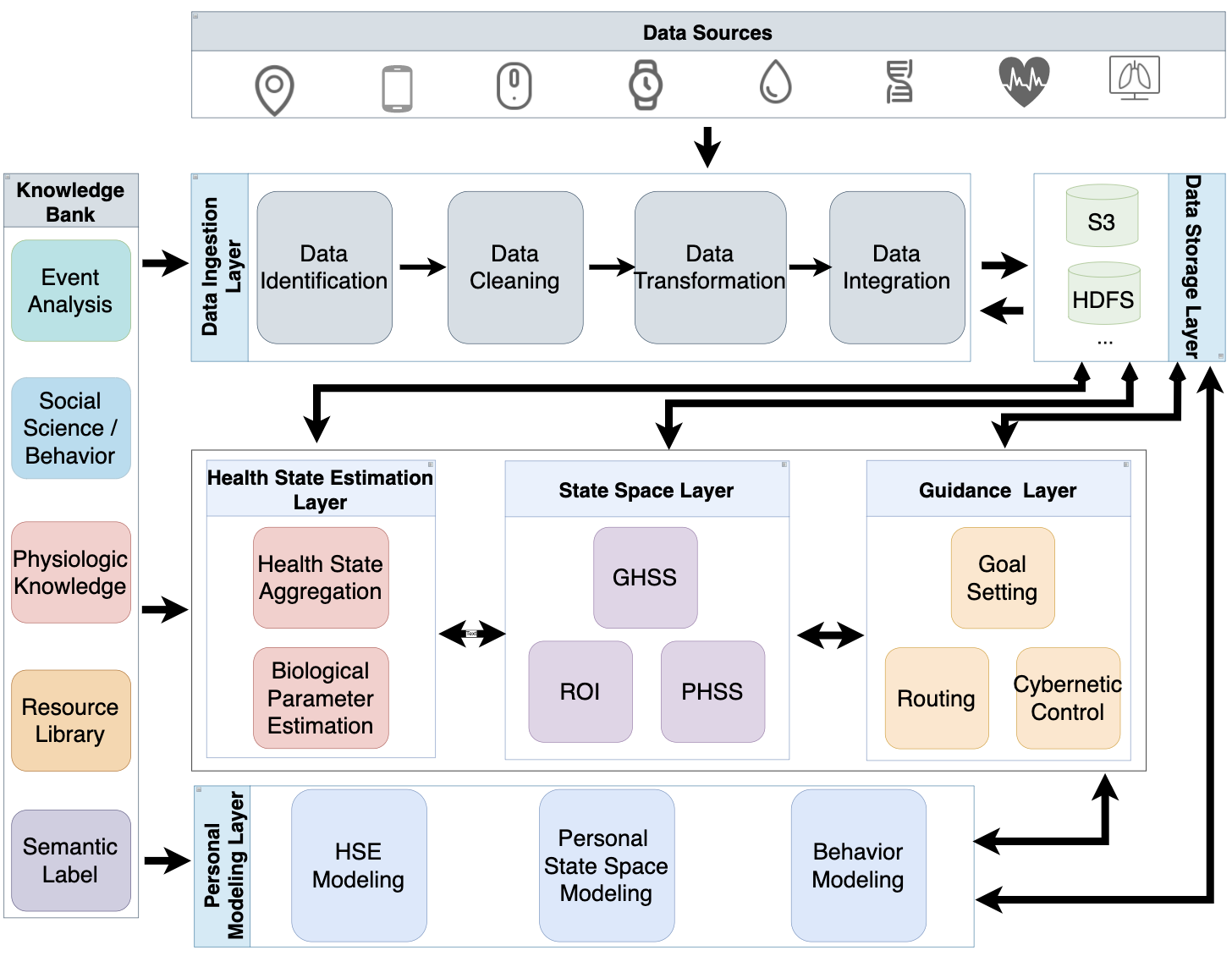}
\caption{General PHN System Framework}~\label{fig:general-system}
  \vspace{-5mm}
\end{figure}

Figure \ref{fig:phnconcept} describes the PHN system from the conceptual points of view. As shown in the Figure, the central purpose of PHN is to guide an individual towards their health goals \cite{Nag2019ALife}. These goals are usually defined computationally in terms of regions-of-interest (ROI) within a multi-dimensional space, where the dimensions represent different components of health as defined through biomedical knowledge. These dimensions build a base map for PHN via the general health state space (GHSS). For a particular user, there is only a subset of this state space they can access due to their biological uniqueness. We call this subset the Personal Health State Space (PHSS). This state space is labeled with ROI that a user may select as a goal. The state space is also populated with the appropriate edges between states, where the edges represent the knowledge of correct inputs to perform a state transition. These concepts can be visually understood in Figure \ref{fig:phnconcept}, step 1. In step 2, we identify where on this map a user exists at the current moment using the appropriate Health State Estimation (HSE) methods. If the user provides a goal, the system can then populate various routes from the current state to the goal state as shown in Figure \ref{fig:phnconcept}, step 3. Upon selection of the desired route, the system transitions to step 4, where control mechanisms are employed to ensure a smooth transition to the next neighboring state along the planned route. All actions carried out (either advised by the system or not) are constantly measured and fed into a new estimation of the state, and controlled to stay on track. This movement of the health state is then used to update the next proposed action. A cycle of these actions ideally produce movement of the user health state closer to the goal health state. Upon reaching the destination, the system continuously ensures that deviation from the goal state is minimized. We propose a system framework to enable this complete flow as shown in Figure \ref{fig:general-system}. We describe each layer of the system in Section 2.1 to 2.5 and apply the framework to a concrete example in Section 3. 




\subsection{Data Ingestion and Data Storage Layer}
The PHN system starts with pulling data to the data ingestion layer in order to meet data for the goal-driven loop. The data ingestion layer pulls 1) stream data and 2) medical / omics data from different data sources, and 3) expert knowledge from the knowledge bank. It also pulls 4) the updated contexts including new health states from the data storage layer after monitoring each user's response to the PHN guidance. Once this data is ingested, a type of data is identified at the beginning of the ingestion process. The data cleaning process follows after it and then this data is converted into different granularities and forms depending on the purpose of the applications. As one of the integration processes, we fuse these heterogeneous data streams into one high-level form of data, which is an event, and index all the data streams by the event in the database \cite{Oh2017FromChronicles}. This could allow for modeling a person in the context of managing a specific aspect of her life using predictive approaches to provide recommendations to maximize the quality of experiences.


\subsection{Health State Estimation Layer}
It is important to note that there is constant flux within an individual's PHSS based on one's location in the state space. For success, navigation must be able to assign an accurate location within the PHSS. Understanding this location is the health state estimation (HSE), which requires ingesting the latest data from the data storage layer and domain knowledge from the knowledge bank to arrive at a predicted current state. For instance, in the case of a cardiovascular health state estimation, wearable sensor data, such as heart rate, activity, step, comes from the data storage layer and the ways of estimating the cardiovascular disease (CVD), such as relative mortality risk using resting heart rate, $VO_2$ Max, Power to Heart Rate Ratio, come from the knowledge bank \cite{cooney2010elevated}. The HSE reports both a location and an accuracy range. The accuracy of the HSE is important to consider as different applications require different levels of accuracy in order to provide services to the user. The same HSE tool may be useful for many applications. For example, monitoring a cardiovascular health state is useful to both endurance athletes and heart disease patients. Estimation techniques have been of great interest in designing many applications, but health applications will require an increasingly deep biological knowledge bank to define and refine the estimated health states that are computed from incoming data. Finally, the estimated health states are shared with the state space layer and stored in the data storage layer.

\subsection{State Space Layer}

The value of good health to an individual is largely based upon how they wish to live. For navigation to occur, first of all, either setting a goal or specifying domain of interest should be preceded by the user in the guidance layer. When the goal is provided by the guidance layer, the system will need to retrieve the appropriate state space given the semantic goal in the state space layer. This process includes identifying the unique set of dimensions, which are consist of health states estimated by the HSE layer, relevant for the specified goal. Goals are essentially states that can be specified as a ROI within these navigational dimensions. The GHSS describes the maximal size state space in which a human can exist.  For example, if the state space of interest is the cardiovascular state, all possible estimations for the cardiovascular state can be considered, including all components of fitness, cardiovascular diseases, structural formation and more. This state space is then further refined for each individual into a PHSS based on individual characteristics (i.e. genetics, gender, age) specific to the individual that provide boundary thresholds. These dimensions are converted into discrete nodes as in Figure \ref{fig:phnconcept}, and then connected via input knowledge that can be knowledge-driven at cold-start and then iteratively improved with data-driven analysis. This PHSS infused by ROI is shared with the guidance layer, HSE layer, and personal modeling layer.


\subsection{Personal Modeling Layer}
Within the PHSS, there are connections/edges made between each of the individual state possibilities as in Figure \ref{fig:phnconcept}. Each edge within the network represents transitions that take the person from one node to another. The state transition takes into consideration all the inputs that will lead to the next state, therefore making connections or relationships between states. Personal modeling is the transformation of the inputs to the predicted output. In the personal modeling layer, the PHN system discovers these relationships in the data and predicts how a certain input should affect the current health state by extending the model into a future time point.

The personal model can include various types of relationship mapping. The HSE modeling is to understand precisely how specific inputs cause an effect on the health state of an individual. Inputs can vary from lifestyle choices, medicine, environment, and more. Biologically, the inputs cause a change in metabolism and gene expression in the user's cells which then change the structure and function of tissue in the organs \cite{Chinsomboon2009TheMuscle}. This change of biological architecture is reflected in a changed health state. The personal state space modeling is to understand the PHSS in more detail. It requires identifying the knowledge layers and ROI within the space relevant to the topic of interest. 

As an initial step, in this paper, we map these relationships with known domain knowledge in the personal modeling layer. The domain knowledge is transferred to a rule-based algorithm that represents the most current understanding of biomedical sciences. Once each individual user accumulates enough data, we plan to build the personal model more precisely by modifying the base personal model by matching the user's patterns to data-driven clustering of user sub-groups. The final stage of personalization would be to modify this sub-group with data produced only through the individual.


To understand the PHSS in more detail, we must identify the knowledge layers and ROI within the space relevant to the topic of interest. Connections within a PHSS between two nodes will be unique based on the individual. For example, for person A and B to improve heart health parameters, the inputs needed for A may be different from person B (This modeling is demonstrated in our experimental section). Therefore the state transition network is also a unique layer on top of the PHSS. Personal models may use various levels of specificity, such as grouping as a sub-population when data is readily available from a single user, as is common in cold start scenarios. Additionally, modelling clustering using traces of health states and state transitions and developing machine learning models based on these clusters may lead to improved personalization while speeding-up the learning time.
 

\subsection{Guidance Layer}


\subsubsection{Route Planning}
After measuring, estimating, modeling the individual, and receiving a goal the user needs to receive guidance on the next step needed to reach that goal. Having a map, location, and goal, in the guidance layer, now sets the stage for routing from the current state towards the goal state as shown in Figure \ref{fig:phnconcept}. Making a route on a map requires not only knowing the start and endpoints but also connecting the two. PHN computes the set of intermediate states and sub-goals with its layer of information, which is relevant for mapping, along with costs and constraints to transition between intermediate states along the planned route.Problem-solving techniques, along with appropriate routing algorithms, reveal the best intermediate states for the user to reach the goal. There may be multiple routes to get to the desired goal. However, route selection could be made by various optimization criteria that include user preferences, efficiency, speed, and available resources.

\subsubsection{Cybernetic Control}
Cybernetic control pairs the individual user and digital assistance to enact real-world actions to transition the health state. Control mechanisms steer individuals towards the state transition on various timescales. At each moment, contextual and timely instructions are given for the next appropriate actions. Actionable inputs that can be part of the guidance include lifestyle events (exercise, nutrition, meditation, etc.) and medical events (medications, procedures, etc.) as examples. Recommendations can serve as a tool for the cybernetic control in a particular moment in time. Recommendation may help plan for a single moment in time, but it does not consider routing through the state space, hence recommendations alone are insufficient. This is a key difference between navigation and recommendation. Control can also be obtained through automatic actuation of inputs, such as changing the thermostat and lighting in the home and screen devices to automatically help re-adjust circadian rhythm from jet-lag \cite{f.lux2021F.lux:Better,Nagare2019DoesSuppression,Sack2009TheLag}. A PHN developer must also consider minimum data requirements (accuracy, sampling frequency etc.) to provide effective control in a given setting. This control can be described in the following equations:
\vspace{-3mm}

\begin{equation}  
    X[k+1] = A[k]X[k] + B[k]U[k]
      \vspace{-3mm}
\end{equation}

\begin{equation}  
    Y[k] = C[k]X[k] + D[k]U[k]
\end{equation} where \textit{X}, \textit{U}, and \textit{Y} are the system true state, inputs, and measured output vectors respectively. \textit{A}, \textit{B}, \textit{C}, and \textit{D} are matrices that provide the appropriate transformation of these variables at a given time \textit{k}. Human health can be described by a state system, and the previous state and the inputs into the system play a role in determining health at time \textit{k+1}.


\section{PHN Experiments in Cardiology}
We applied the generalized PHN framework into a specific application to test the performance of PHN in the real world. We set the health state space dimensions to represent cardiovascular health and cardio-respiratory fitness (CRF) of each of the PHN users, with the goal set to improve these two global dimensions towards the PHSS optimal state for a given user. Cardiac health is the most significant cause of death in humans. However, in healthcare systems today, it is more commonly addressed in high need or critical situations using expensive lab tests and rehabilitation programs \cite{Ross2016ImportanceAssociation}. 

To tackle this challenge, in our first experiment, we tested the use of commercial wearable devices deployed with the PHN framework to help users improve their cardiac health state. In this experiment, as an initial effort of building the navigation system, we built a rule-based HSE model using domain knowledge derived from bioenergetics science. With this model, we provided actionable daily exercise guidance through the cardiac PHN system to 41 participants and monitored the changes in their CRF indicators with additional testing during the experiment. In the second experiment, we built a data-driven personal model to test the personal state space modeling on cardiac exercise response variability from person to person. In this experiment, we aimed to emphasize the necessity of building the personal model by showing that a standard dose of exercise would not effectively impact each participant's cardiovascular health state in the same way \cite{ross2019precision}. To do this, we developed a simple classification model for cardiac exercise response variability through a minimum feature set. We then saw to what extent this approach can allow for rapid cold-start personalization. 
Developing the advanced personal models using state-of-the-art algorithms would be handled in the future once real world data is sufficiently collected.






\subsection{Experiment 1: Prospective Clinical Trial}
We implemented a knowledge-infused cardiovascular PHN system that aggregates multi-modal sensor data streams from each participant's smartwatch and mobile device, estimates their current heart status, computes the best route to achieve the health goal, and provide actionable daily guidance and control to participants. The closed-loop PHN system captured the participants' reactions to the guidance, and dynamically modified the next iteration of guidance. Our aim in this approach was to make improvement in each participant's cardiac health state and observe changes in their lifestyle, while still allowing for the stochastic environment of normal daily life.

\subsubsection{PHN Implementation for Cardiovascular Health Navigation}

\begin{figure}
\small
\centering
\includegraphics[width=.8\columnwidth]{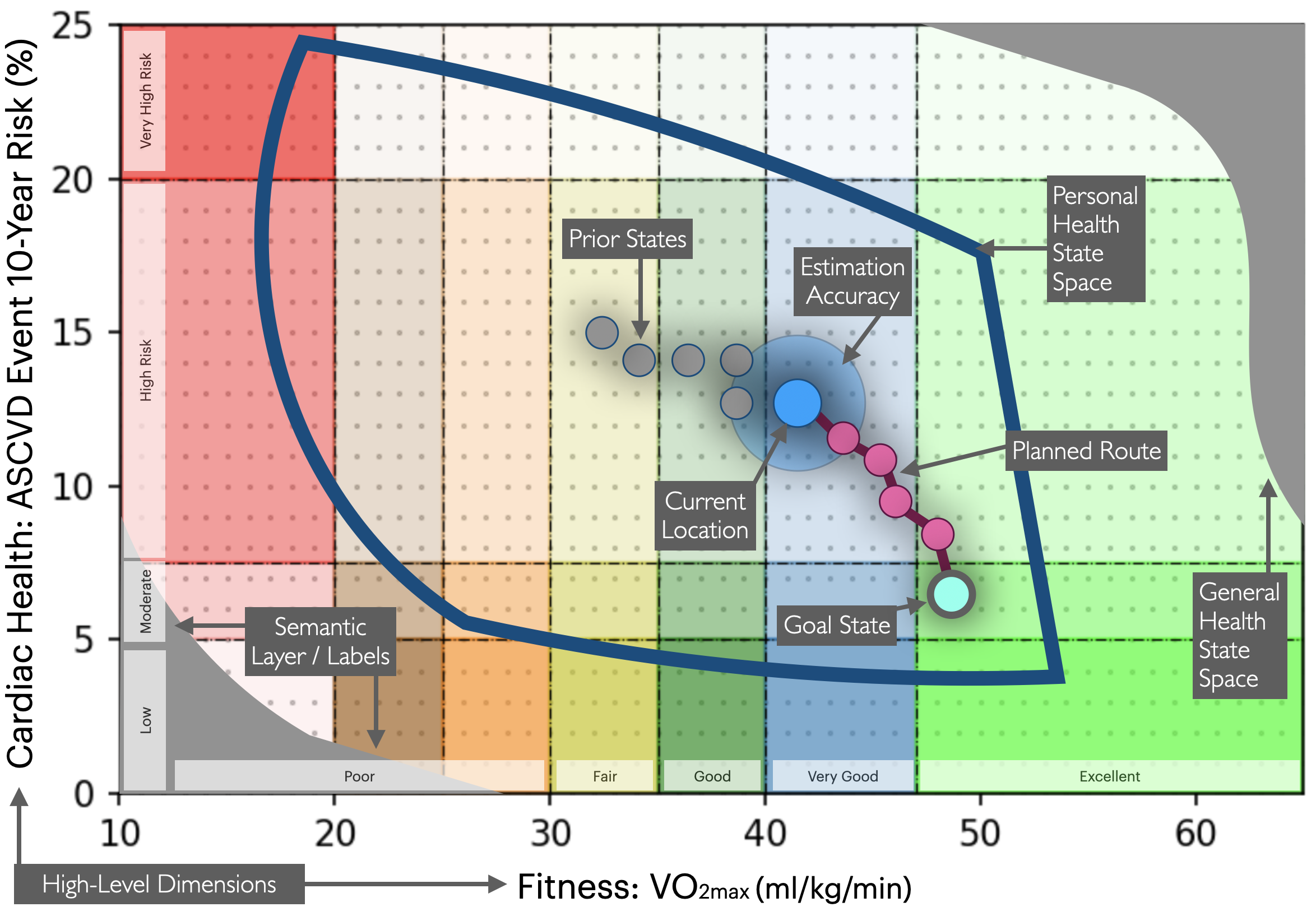}
\caption{The Heart Health Navigation State Space. Regions of Interest are described in different colors along with their semantic labels.}~\label{fig:phss}
\end{figure}

Based on Figure \ref{fig:general-system}, we built an ETL\footnote{Extract Transform Load} pipeline on AWS\footnote{Amazon Web Service} to do batch processing for each participant's data sources in a database, then processed the data in each layer using server-less functions (Step, Lambda) on AWS to arrive at the daily guidance.

\textbf{Data Ingestion and Data Storage Layer}\label{lab:section:3_1_1}. We ingested the following data with data cleaning, transformation and integration and stored in S3 bucket on AWS:
\begin{itemize}
    \item Expert Knowledge: Sports science, bioenergetics science.
    \item Stream Data: time stamp, step count, HR (BPM), activity mode (e.g., still, walking, running), sleep (e.g., deep sleep, light sleep, rem sleep, sleep score), resting HR, age, gender, height, weight.
    \item Medical Data: ASCVD \cite{Henderson2016VALIDATIONSTUDY}\footnote{Atherosclerotic Cardiovascular Disease, which is measured by cholesterol levels, diabetes status, smoking habits, blood pressures, age, and gender.}, cardiac risk factors.
\end{itemize}
\textbf{Health State Estimation Layer}. To estimate the cardiovascular health state, we calculated cardiac disease risk by weighting the user's ASCVD risk and further enhanced with relative risk modifier of resting heart rate extracted during deep sleep demonstrating the combination of both episodic high quality medical blood data with high frequency sensing of a wearable device. These health state techniques were pulled from the knowledge layer \cite{Stone20142013Adults, diaz2005long, cooney2010elevated, jensen2013elevated}. Every week we also measured indicators of maximum oxygen consumption capacity (VO$_{2}$Max) through two exercise tests, a 3-minute step test and 6-minute walking test \cite{bohannon2015six}. With these indicators, we estimated and monitored each participant's current heart health state.




\textbf{State Space Layer}. With these two dimensions (ASCVD, VO$_{2}$Max) estimated by the HSE layer, we generated a GHSS heart map for all the users at the beginning of the trial. This demonstrates the important integration of both high quality episodic medical data such as blood laboratory values combined with high frequency data collected from wearable devices to estimate fitness status. We then applied knowledge about cardiovascular health \cite{Stone20142013Adults,Khuwaja1994ArchitectureTutor} into the map, and specified the ROI. After that, we converted this map to PHSS given the user demographic information (e.g., age, gender) and domain knowledge \cite{ Roth2015DemographicMortality} as shown in Figure \ref{fig:phss}.

\begin{table}
\small
  \caption{Definitions in the guidance module}
  \label{tab:terms}
  \begin{tabularx}{230pt}{cX}
    \toprule
    Term & Description\\
    \midrule
    TRIMP & (TRaining IMPulse) Weighted product of training volume and training intensity by HR.\\
    CTL & Chronic Training Load (Fitness Level). Average TRIMP score for the recent 6 weeks.\\
    ATL & Acute Training Load (Fatigue Level). Average TRIMP score for the recent 1 week.\\
    TSB & Training Stress Balance. CTL $-$ ATL.\\
    TRIMP$_{min}$ & Minimum Weekly Goal based on CDC guideline \cite{piercy2018physical}. \\
    TRIMP$_{d}$ & Daily TRIMP goal \\
    TRIMP$_{w}$ & Weekly TRIMP goal \\
    R & Ramp Rate (\(0 \leq R \leq 1\)) \\
    C$_1$ & Optimal Training Zone Coefficient (\(5 \leq C_1 \leq 30\)) \\ 
    t & Current day \\ 
    \bottomrule
  \end{tabularx}
\end{table} 

\textbf{Personal Modeling Layer: } To build a personal module for the cardiac PHN system, we retrieve from the knowledge layer about how increasing intensity and duration of exercise is lowers risk of cardiovascular disease \cite{Ross2016ImportanceAssociation}. We further extract from the layer advanced physiologic cardiovascular endurance training strategies in bioenergetics science \cite{foster2001new, esteve2007impact, anta2011training, banister1999training, calvert1976systems, simmonswebsite} to build a personalized rule-based model for daily exercise guidance. Table \ref{tab:terms} explains the definitions used in the rule-based model and the following rules are the key rules we used in the guidance module: 

\begin{itemize}
\item TSB \( \geq\) +10: Transition Zone. User is well rested. The value is often reached when a user has an extended break.
\item +5 \( \leq \) TSB \( <\) +10: Fresh Zone. Zone reached when user is optimally recovered from exercise. 
\item -5 \( \leq \) TSB \( <\) +5: Neutral Zone. Zone reached typically when an user is in rest or recovery week.
\item -30 \( \leq \) TSB \( <\) -5: Optimal Training Zone. Zone where the user can best build their effective fitness.
\item -30 \( > \) TSB: Over Load Zone. User is over-training and should take a rest to protect from injury.
\item TSB should be maintained in the optimal training zone to improve ASCVD and VO$_{2}$Max.
\begin{equation}
TRIMP_{w} = CTL_{t-1} \times (1 + R) + C_{1}
\end{equation}
\item CTL increase with a maximal rate limit not exceeding 5 per week.
\begin{equation}
CTL_{t-1} - CTL_{t-8} < 5
\end{equation}
\item TSB shouldn't drop below -20 more than once in 10 days.
\item If the TSB dropped below -20 in a given week, the next week's training goal should be scaled down slightly. 
\end{itemize} 

\begin{figure}
\small
\centering
\includegraphics[width=0.85 \columnwidth]{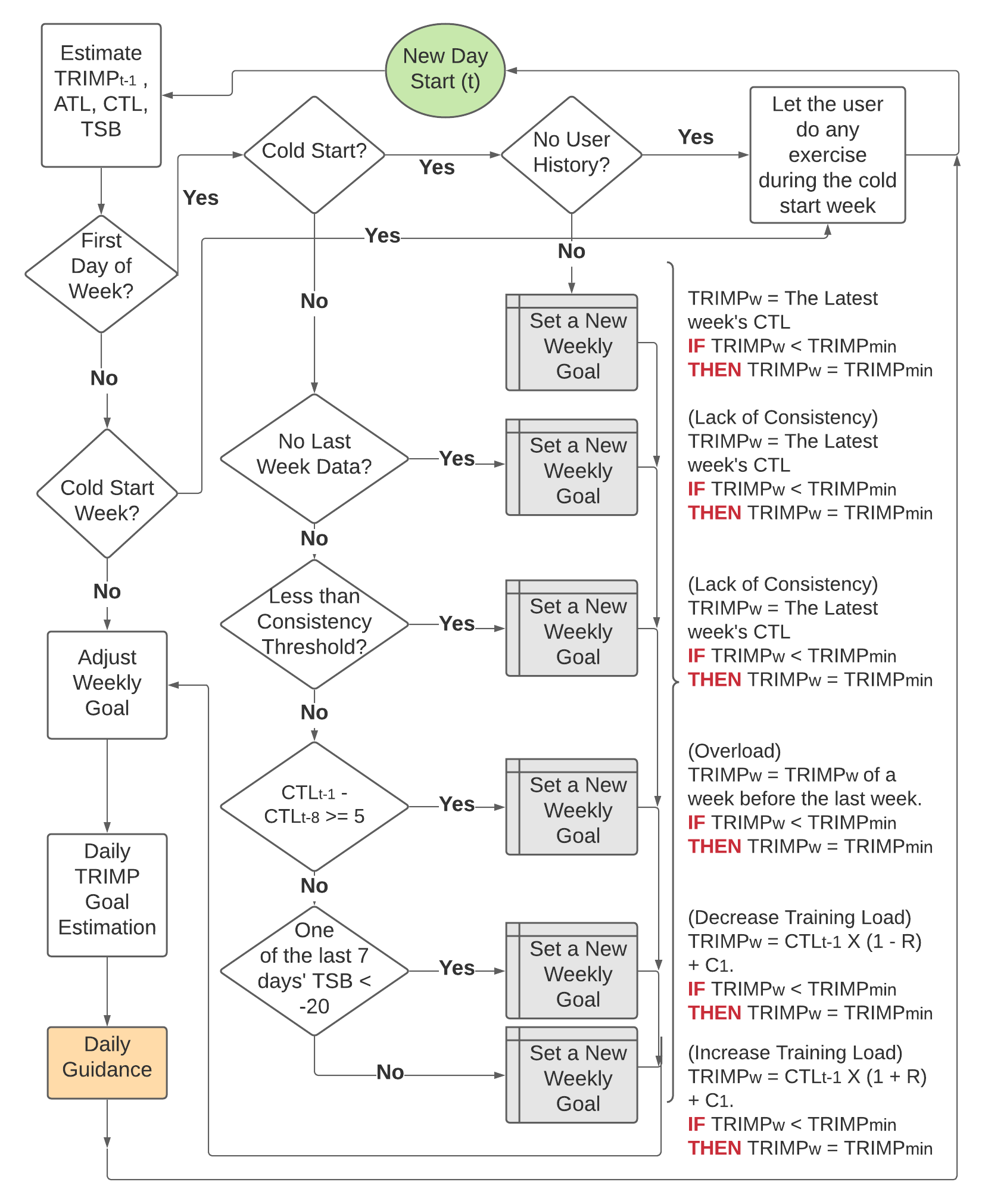}
\caption{A flowchart for the Knowledge-infused daily exercise guidance algorithm .}~\label{fig:rule}
\end{figure}

\begin{figure}
\small
\centering
\includegraphics[width=1.0 \columnwidth]{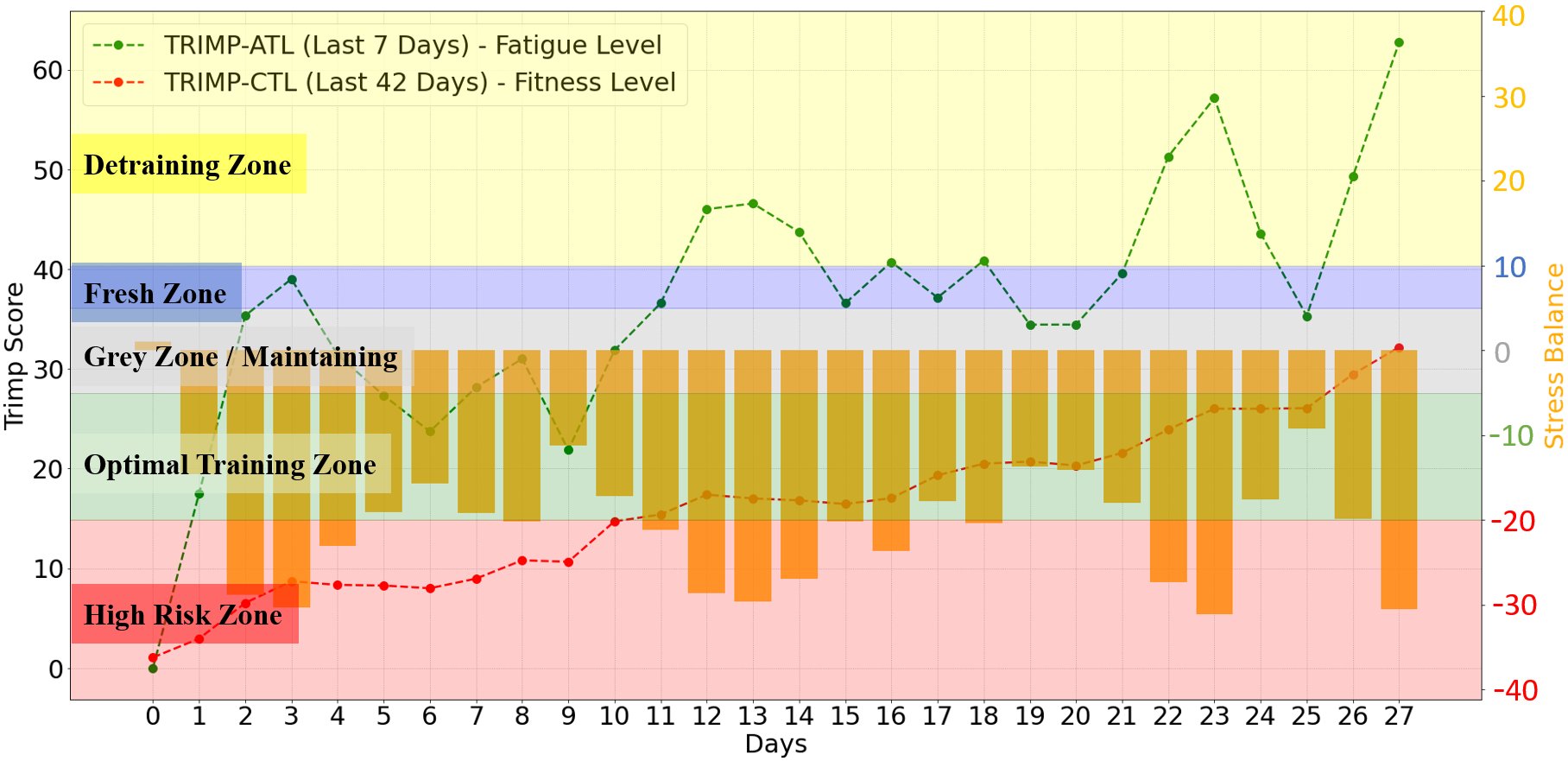}
\caption{Experimental user trend of fitness level (CTL), fatigue level (ATL), and training stress balance (TSB). The user is able to monitor keeping their lifestyle on track by keeping the yellow bars within the green TSB level.}~\label{fig:training}
\end{figure}

\textbf{Guidance Layers}. With the rule-based model we defined in the personal modeling layer, we built a guidance module including routing and control. Figure \ref{fig:rule} shows the flow of how the computations are progressing. By using this guidance module, we tried to maintain the participants within their optimal training zone as shown in Figure \ref{fig:training}, which can potentially help improve the levels of CRF. More importantly, we provided a triplet for daily exercise guidance including exercise type, exercise intensity, and exercise duration, such as Go jogging for at least 40 minutes sustaining HR above 113 bpm. To do this, we converted a daily TRIMP goal into a recommendation triplet as follows: 
\begin{itemize}
    \item Low-intensity workout for $\frac{TRIMP_{d}}{c_1}$ minutes while sustaining $0.55 \times MaxHR  <= HR < 0.70 \times MaxHR$
    \item Medium-intensity workout for $\frac{TRIMP_{d}}{c_2}$ minutes while sustaining $0.70 \times MaxHR  <= HR < 0.80 \times MaxHR$
    \item High-intensity workout for $\frac{TRIMP_{d}}{c_3}$ minutes while sustaining $0.80 \times MaxHR  <= HR <= 1.00 \times MaxHR$
\end{itemize} where $c_1, c_2,$ and $c_3$ are the Lucia's coefficients, which are 1, 2, and 3, respectively \cite{lucia1999heart}. With all of the rules in the guidance module, we gradually found the proper intensity and duration of exercise over time with the cybernetic control in the PHN system. Additionally, since $TSB$ can reflect each participant's current fatigue level and fitness level, we could reflect the current states of the users and provide the adjusted guidance in the cybernetic loop even if the participants did not follow the PHN guidance perfectly. Figure \ref{fig:training} shows that the PHN system is gradually improving the user's $ATL$ and $CTL$ while maintaining $TSB$ in the green zone.
 




\subsubsection{Experimental Setup}
We tested the PHN system with 41 recruited participants for 4 weeks with an Amazfit smartwatch\footnote{https://www.amazfit.com/en/index}. Participant's smartphones were synchronized daily with this data to the PHN cloud. During the trial period, we performed a gold-standard physical test once a week in a place where we measured each participant's blood pressures, weight, and step testing. For daily guidance, we split the participants (N=41) into the PHN treatment group (N=26) and the control group (N=15). This produced 40,320 samples per user and 1,653,120 samples total. For the PHN group, we provided exercise lifestyle guidance including intensity and duration of exercises. We also allowed the participants not to follow the guidance if they felt exhausted or achy since the PHN can reflect the participants' current behavior change and adjust the intensity and duration dynamically at the personal level. We gave the control group freedom and did not provide daily exercise guidance, other than gold-standard features of daily exercise goal tracking in Amazfit (e.g., step goals, active minutes etc.).

\subsubsection{Experiment Results}

\begin{table}
\small
   \caption{PHN groups improved in their Systolic and Diastolic Blood Pressure, BMI, VO$_{2}$Max Indicators, and HR Recovery. The mean and standard deviation of each indicator was measured at the start and end of the experiment. The difference between these two values and their p-value was also measured in the end.}
  \label{tab:e2results}
  \begin{tabularx}{445pt}{rXXXXXX}
  \toprule
  \multirow{2}{*}{Indicator} & \multicolumn{3}{c}{Treatment Group (N=26)} & \multicolumn{3}{c}{Control Group (N=15)} \\
  
                           & \multicolumn{1}{c}{Start (Mean, SD)}         & \multicolumn{1}{c}{End (Mean, SD)}        & \multicolumn{1}{c}{Change}       & \multicolumn{1}{c}{Start (Mean, SD)}        & \multicolumn{1}{c}{End (Mean, SD)}         & \multicolumn{1}{c}{Change}         \\
    \midrule
\multirow{2}{*}{Systolic BP}         & \multicolumn{1}{c}{119.69}  & \multicolumn{1}{c}{114.00}  & \multicolumn{1}{c}{-4.76}  &  \multicolumn{1}{c}{120.6} & \multicolumn{1}{c}{116.07} &   \multicolumn{1}{c}{-3.76}        \\
                           & \multicolumn{1}{c}{(11.55)}  & \multicolumn{1}{c}{(12.19)} & \multicolumn{1}{c}{(\textbf{p < 0.001})}   &   \multicolumn{1}{c}{(11.87)}        &   \multicolumn{1}{c}{(10.30)}        &  \multicolumn{1}{c}{(p = 0.061)}         \\
\multirow{2}{*}{Diastolic BP}         & \multicolumn{1}{c}{75.19}  & \multicolumn{1}{c}{72.35}  & \multicolumn{1}{c}{-3.79}  &  \multicolumn{1}{c}{76.47} & \multicolumn{1}{c}{76.27} &   \multicolumn{1}{c}{-0.26}        \\
                           & \multicolumn{1}{c}{(8.98)}  & \multicolumn{1}{c}{(9.07)} & \multicolumn{1}{c}{(\textbf{p = 0.009})}   &   \multicolumn{1}{c}{(6.36)}        &   \multicolumn{1}{c}{(5.34)}        &  \multicolumn{1}{c}{(p = 0.862)}         \\
\multirow{2}{*}{BMI}         & \multicolumn{1}{c}{24.03}  & \multicolumn{1}{c}{23.73}  & \multicolumn{1}{c}{$-1.25$}  &  \multicolumn{1}{c}{24.54} & \multicolumn{1}{c}{24.51} &   \multicolumn{1}{c}{-0.11}        \\
                           & \multicolumn{1}{c}{(3.12)}  & \multicolumn{1}{c}{(3.00)} & \multicolumn{1}{c}{(\textbf{p = 0.009})}   &   \multicolumn{1}{c}{(2.06)}        &   \multicolumn{1}{c}{(2.06)}        &  \multicolumn{1}{c}{(p = 0.838)}         \\
\multirow{2}{*}{Step Testing}         & \multicolumn{1}{c}{46.18}  & \multicolumn{1}{c}{49.41}  & \multicolumn{1}{c}{+7.00}  &  \multicolumn{1}{c}{46.95} & \multicolumn{1}{c}{49.53} &   \multicolumn{1}{c}{+5.50}        \\
                           & \multicolumn{1}{c}{(8.68)}  & \multicolumn{1}{c}{(10.47)} & \multicolumn{1}{c}{(\textbf{p = 0.014})}   &   \multicolumn{1}{c}{(8.29)}        &   \multicolumn{1}{c}{(8.01)}        &  \multicolumn{1}{c}{(\textbf{p = 0.013})}         \\
\multirow{2}{*}{HR Recovery}         & \multicolumn{1}{c}{121.10}  & \multicolumn{1}{c}{113.32}  & \multicolumn{1}{c}{-6.42}  &  \multicolumn{1}{c}{117.99} & \multicolumn{1}{c}{111.05} &   \multicolumn{1}{c}{-5.88}        \\
                           & \multicolumn{1}{c}{(17.18)}  & \multicolumn{1}{c}{(13.38)} & \multicolumn{1}{c}{(\textbf{p = 0.005})}   &   \multicolumn{1}{c}{(12.64)}        &   \multicolumn{1}{c}{(12.52)}        &  \multicolumn{1}{c}{(\textbf{p < 0.026})}         \\

    \bottomrule
  \end{tabularx}
\end{table} 

We evaluated the PHN system with two analysis perspectives, objective and subjective. For the objective analysis, we examined the CRF indicators in Table \ref{tab:e2results} of each participant every week and compared the changes of the indicators between the PHN group and the control group. The PHN group significantly (paired t-test p<0.05) improved 5 CRF indicators and the improvement outweighed the control group. For systolic blood pressure, the PHN group decreased 4.76\% on average, versus the control group with no significant change. For diastolic blood pressure, the PHN group decreased 3.79\% on average, versus the control group had no significant change. For BMI, the PHN group decreased 1.25\% on average, versus the control group had no significant change. For VO$_{2}$Max estimated by the step test and walking test, the PHN group increased 7\% on average versus the control group increased 5.5\%, showing PHN was 27.3\% better at improving VO$_{2}$Max indicators. Within HR recovery, a lower average HR indicates a better speed of recovery, the PHN group decreased 6.42\% on average, versus the control group decreased 5.88\%, showing PHN was 9.2\% better at improving HR recovery. These results show that the rules we extracted from expert/medical knowledge can better help navigate the participants to achieve their goal compared to those of the participants who exercised with freedom. It can also indicate that the daily goals and weekly goals based on the participants' current states effectively act as the intermediate states along the planned route. 

\begin{table}
\small
  \caption{PHN Users Describing their Experiences}
  \label{tab:e3results}
  \begin{tabularx}{240pt}{Xcc}
    \toprule
    Survey Questions & Number of Users & Score [0, 10] \\
    \midrule
    Q1. \textit{How satisfied are you with using PHN?} & 38 & 8.55 (SD=1.13) \\ \\
    Q2. \textit{How satisfied are you with your health/exercise improvement using PHN?} & 38 & 7.47 (SD=2.17) \\ 
    \\
    \Xhline{2\arrayrulewidth}
    
    Survey Questions & Number of Users & Percentage(\%) \\
    \midrule
    Q3. \textit{Did PHN help you make better exercise habits?} & \begin{tabular}{@{}c@{}}31 (Yes) \\ 7 (No)\end{tabular} & \begin{tabular}{@{}c@{}}\textbf{81.58 (Yes)} \\ 18.42 (No)\end{tabular} \\
    Q4. \textit{Did following PHN reduce your perceived risk of injury?} & \begin{tabular}{@{}c@{}}29 (Yes) \\ 9 (No)\end{tabular} & \begin{tabular}{@{}c@{}}\textbf{76.32 (Yes)} \\ 23.68 (No)\end{tabular} \\

    \bottomrule
  \end{tabularx}
\end{table}

Subjective analysis via exit interview with the participants revealed an average score of 8.55 (rating from 0 to 10) to the guidance of the PHN system and users confirmed that the PHN system has improved their well-being with an average score of 7.47/10. 81.58\% participants confirmed that the PHN system helped establish exercise habits, 76.32\% participants thought the PHN system helped prevent them from injuries, and 84.21\% participants felt they improved their CRF by the PHN system as shown in Table \ref{tab:e3results}. These results show that PHN can help maintain the users' engagement in exercise while improving their health states. This result shows promise towards inspiring long-term adoption of healthy habits\cite{clawson2015no}.






In summary, experiment 1 demonstrates that PHN can be a powerful adjunct or alternative therapy for treating and preventing chronic cardiovascular disease.




\subsection{Experiment 2: Prediction of Exercise Response Variability}

 

In this experiment, we focused on showing that the connections within a PHSS between two nodes will be unique for a person through a data-driven personal model describing the best inputs for the state change. To do this, we first combined, processed, and analyzed user exercise, heart rate and sleep data and showed there are clustering in user CRF responses. We then applied the cluster information matched to a user CRF response type to personalize daily guidance for users. Lastly, we use individual data to model personal response lag time for predicting state transition on the personal state space.

\subsubsection{Exercise Feature Categorization} In a situation of similar exercise program dose, we defined exercise group based on frequency (Low: 1 time per week, Mid: 2 - 4 times per week, High: 5 - 7 times per week), amount (Low: less than 30 minutes per exercise, High: more than 30 minutes per exercise), and intensity of exercise (Low: lower than 75 \% of estimated maximal heart rate,  High: greater than 75 \% of estimated maximal heart rate) \cite{ross2019precision}. We then categorized each individual into 12 different exercise groups, such as High (Frequency) - High (Amount) - High (Intensity).




\subsubsection{Labeling for classification} 
Strong causal relationships connect CRF to predicting cardiovascular disease \cite{bouchard1994physical}. Additionally, epidemiological research shows that resting heart rate is an independent predictor of cardiovascular disease \cite{fox2007resting}. Given this evidence, we utilized resting heart rate as an indicator of CRF and labeled each individual's cardiovascular exercise response as positive ($V_{rhr}(t) <   -0.5 $), neutral ($-0.5 <= V_{rhr}(t) <=  0.5$), and negative responder ($V_{rhr}(t) >  0.5$) by the velocity of resting HR change 
\begin{equation}
V_{rhr}(t) = \frac{\sum_{i=1}^{t}(x_i - \bar{X})(y_i - \bar{Y})}{\sum_{i=1}^{t}(x_i - \bar{X})^2},
\end{equation}
where t is total weeks, $x_i$ is 1,2,3,...,t, $y_i$ is the resting heart rate of $i_{th}$ week, $\bar{X}$ is the sum of $x_i$, and $\bar{Y}$ is the sum of $y_i$ \cite{ross2019precision}. The positive responder means a person who got decreased his resting heart rate after certain weeks of exercise. The negative responder means a person who got increased his resting heart rate after certain weeks of exercise. The neutral responder means a person who didn't have a big change in his resting heart rate even after certain weeks of exercise.


\subsubsection{Data set}
We used 33,269 anonymous Amazfit users' smartwatch sensor data which satisfies the following criteria:
\begin{itemize}
    \item Criteria 1: The users should have been continuously using their smartwatch for the recent 3 months.
    \item Criteria 2: The quality of sensor data should be in a reasonable range (e.g., continuity > 70\%, accuracy > 70\%).
    \item Criteria 3: The users should have at least one exercise event per week for 4 consecutive weeks.
\end{itemize} 
We used the same kinds of smartwatch sensor data streams as the ingested data in Experiment 1. We processed ~10 billion minute level samples. From heart rate, activity mode, and GPS data we obtained exercise-related features and sleep-related features as follows:
\begin{itemize}
    \item Exercise-related Features: the number of exercise per week, average weekly exercise time, average weekly exercise heart rate, average weekly active time.
    \item Sleep-related Features: average weekly sleep score, average weekly sleep time and its stages (e.g., deep sleep, light sleep, rem sleep, awaken time), average weekly number of wake ups during sleep.
\end{itemize} 


\subsubsection{Selecting Experiment Data Set}
\begin{figure}
\small
\centering
\includegraphics[width=1 \columnwidth]{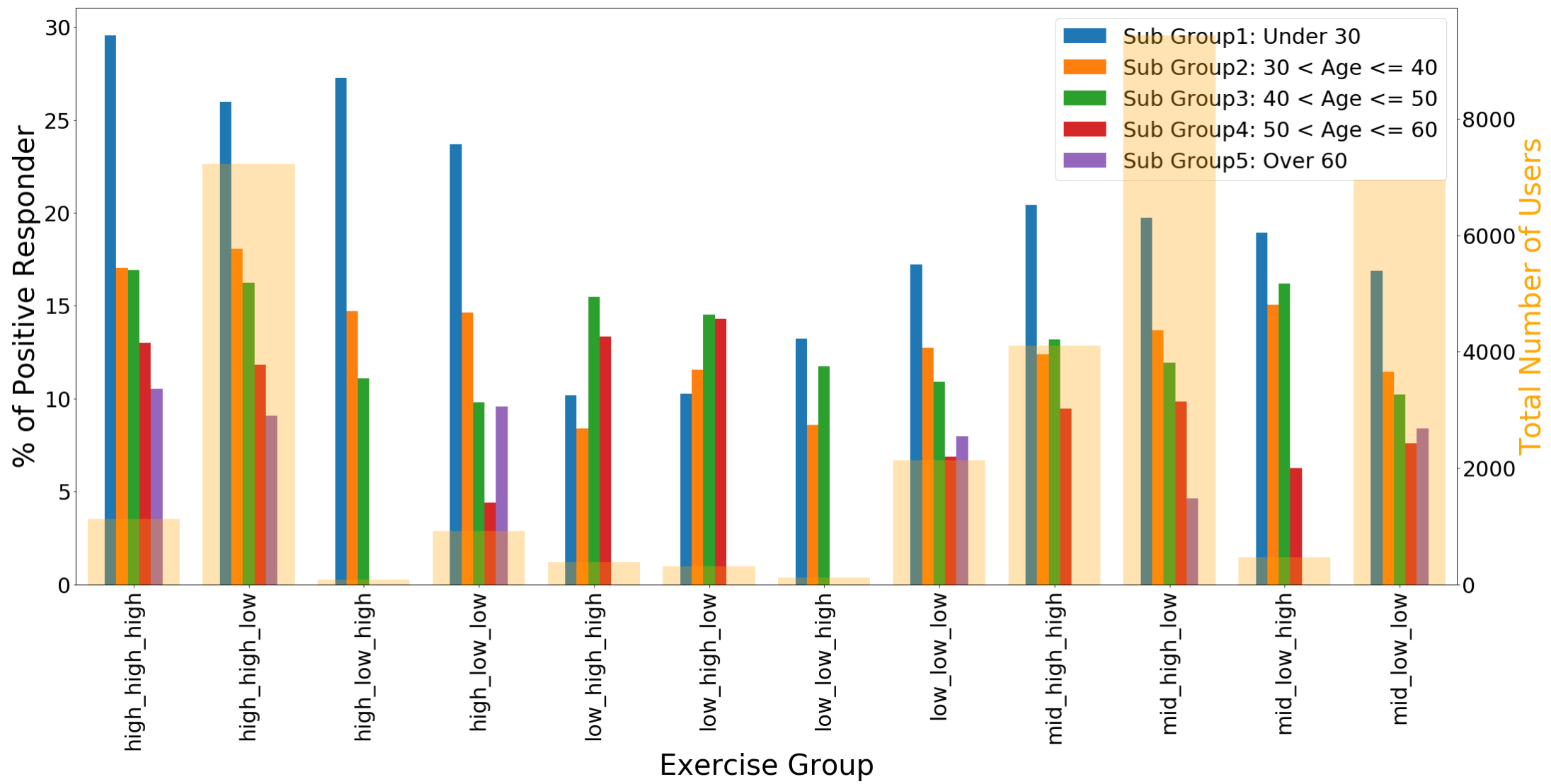}
\caption{Variation in exercise/age group on response. X-axis labels refer to Frequency, Volume, Intensity).}~\label{fig:select_exercise_group}
\end{figure}

Different frequencies, amounts, and intensity demonstrate unique personal responses to exercise as shown in Figure. \ref{fig:select_exercise_group}. Additionally, not all age groups react the same to exercise. These observations indicate that some groups are higher in exercise response variability than the others, which suggests the need for personalization. To dive deeper into the personalization, we selected age sub-group 1 (<30 years) in the High-High-Low exercise group where the total number of user response types out of 1,175 (positive: 305, neutral: 787, negative: 83) and studied classification of users into different responder groups.

\subsubsection{Cardiovascular Exercise Response Variability Classification}

\begin{table}
\small
  \caption{Model selection through repeated stratified 10-fold cross validation. The following precision, recall, and F1 score are the weighted results.}
  \label{tab:10_fold}
  \begin{tabular}{rccc}
    \toprule
    Classifier &Precision & Recall & F1 \\
    \midrule
    Neural Networks & \textbf{0.6487}& 0.6276 &0.6379\\
    Logistic Regression & 0.6480& \textbf{0.6992} & \textbf{0.6424} \\
    Support Vector Machine (SVM)& 0.6113& 0.6803 &0.5832\\
    Random Forest& 0.6072& 0.6802 &0.6089\\
    Gradient Boosting Machine (GBM)& 0.6144& 0.6803 &0.5772\\
    XGBoost& 0.6009& 0.6602 &0.6206\\
    \bottomrule
  \end{tabular}
\end{table}

\begin{table}
\small
  \caption{Evaluation for the hyper-parameter tuned Logistic Regression classifier}
  \label{tab:train_test}
  \begin{tabular}{cccccc}
    \toprule
    Label &Precision & Recall & F1 & Support\\
    \midrule
    Negative& 1.0& 0.1053 & 0.1905 &19 \\
    Neutral& 0.7119 & 0.9105 &0.7991 &190\\
    Positive& 0.5135& 0.2603 &0.3455&73\\
    \bottomrule
  \end{tabular}
\end{table}

First, we split the filtered 1,175 users into the train (75\%) / test (25 \%) data sets. We then applied classification models with basic user information (e.g., age, gender, weight, height) on the train data set through a repeated stratified 10-fold cross-validation to see to what extent these basic features, which can be obtained from the users at the beginning, can allow for rapid cold-start personalization. The f1 scores of logistic regression, support vector machine (SVM), random forest, gradient boosting machine (GBM), XGBoost, and neural networks are 0.288, 0.288, 0.331, 0.292, 0.320, and 0.285, respectively. After that, we used the users' 7-days worth of data including the features defined in Section \ref{lab:section:3_1_1}, and repeated the above experiment to see how much the performance can be improved on the first week's data. The results are shown in Table \ref{tab:10_fold}. Based on the results, we used Logistic Regression to evaluate the classifier. We trained the model with the same training data set, tuned the hyperparameters, and tested the model on the same testing data set. The results are shown in Table \ref{tab:train_test}. We can see from these two results an improvement in f1 of 0.4032 through the one-week sensor data, especially using the exercise/sleep-derived features, compared to the model that only used basic user information. We can also see from the results that even the simple classification models can show a reasonable performance once the feature sets are correctly derived.

We believe that building a personal model requires consistently finding inter-individual differences over time and personalizing the model with these factors, such as with federated learning \cite{Li2020FederatedDirections}. We think this approach is especially important as the majority of the people (67\%) do not show standard responsiveness to external influences as shown in Figure \ref{fig:select_exercise_group}. Since it is difficult to know these differences initially, we can use a sub-population based model, which is what we did in this section, if there is not enough data readily available from a single user followed by incrementally reinforcing the model in a closed-loop cycle of the PHN framework.







\section{Conclusions and Future Opportunities}
 This paper stitches together a unified personal health computing framework called Personal Health Navigation and tests the effectiveness of the concept with real-world users. Guidance in PHN blends together the capabilities of using fused MM data with cybernetic control and long-term intelligent planning. We demonstrate the effectiveness of PHN in a pilot clinical trial with promising results, and create big-data based advanced personal health models. We believe this demonstration of PHN sets a key path forward by which advanced systems for personal health computing can be built. 
 
 Modern approaches to health will require a high resolution personalized sense-to-action paradigm that can be active at all times, not just when a person falls ill. This sense-to-action approach with PHN creates a continuous and quantitative view of health that can be applied proactively as an intrinsic part of one's daily decisions, either lifestyle or via medical professionals. Applications of PHN can reach far into many fields of interest beyond traditional health, such as social media, entertainment, sports, travel, education, business, and shopping. As examples, a student would study best when they are mentally and physically best prepared to learn, hotels would be able to personalize room atmosphere based on user thermal and circadian jet-lag status, and athletes would be able to perform their best for the Olympics. We believe that PHN creates further opportunities to the following challenges.

\begin{enumerate}
    \item \textbf{Personalization:} As alluded to in the introduction, personalization through the creation of robust personal models is the centerpiece of a high quality PHN system. This will require investigation into modelling methods across a variety of data types and for different specialities.
    \item \textbf{Human-Computer Interaction (HCI) for Health:} For digital health to be successfully implemented, the user must have a seamless and enjoyable experience for daily use. Digital forms of health interaction are rising rapidly with interactive based exercise, nutrition, and mental health platforms \cite{Peloton2021PelotonOn-Demand,Grubhub2021Grubhub,ModernHealth2021ModernWorkforce,Ginger2021GingerHealthcare}. Advancing this will require understanding the unique aspects of health related HCI.
    \item \textbf{Deep Health Data Interpretation and Summarizing:} Taking actions for health requires understanding of high dimensional space with various types of data. Humans cannot reliably take advantage of large amounts of raw data unless it is quickly interpreted and summarized appropriately for a given audience. Integration of complex medical data along with sensors in daily life, hence will require unique analysis techniques. Aggregating and summarizing this data to be relevant to the general user or health professional remains an open challenge.
    \item \textbf{Indexing:} Intuitive and computationally useful indexing will depend on labeling composite data segments efficiently and semantically understandable. Event detection techniques must take advantage of user produced data through combining multimedia (audio-visual-tactile-gustatory-olfactory) and multimodal (HR, EKG, movement, GPS etc.) signals.
    \item \textbf{Storage and Transport:} With the acceleration in how much data an individual creates, the management and communication of data has increasing need for research in techniques related to digestion, cleaning, synchronization, unification, compression, and retrieval in order to make the data useful \cite{kaempchen2003data}.
    \item \textbf{Knowledge Systems:} Much of the domain knowledge pertaining to health will require extraction and transformation of biomedical knowledge into a computationally usable form. This requires extending traditional knowledge techniques for a higher level of multimodality  \cite{Haussmann2019FoodKG:Recommendation,Zulaika2018EnhancingGraphs,Abu-Salih2020Domain-specificSurvey,Rotmensch2017LearningRecords}.
    \item \textbf{Event Mining:} Mining event streams of various types is essential for enhancing the personal model and understanding contextual situations with greater richness as shown in Fig. \ref{fig:general-system}. Event mining research has the potential to dramatically enhance the computational interpretation of MM data related to personal health. 
    \item \textbf{Recommender Systems:} PHN provides the umbrella by which context-aware health recommendations can be computed, but will require further investigation to improve efficacy. Recommendations must balance user health, resources, preferences, and a host of competing goals. This leads to complex optimization problem solving issues to produce a high performing recommendation engine. Contextual accuracy and precision alone have been shown to improve recommender effectiveness, hence the need for deeper research to find further avenues for improvement \cite{Lin2011Motivate:Xplore,Patel2015WearableChange}.
\end{enumerate}

\bibliographystyle{ACM-Reference-Format}
\bibliography{sample-base}

\end{document}